\documentclass{article}
\usepackage{iclr2027_conference,times}

\usepackage{hyperref}
\usepackage{url}

\usepackage{pifont}
\usepackage{amsmath,amssymb}
\usepackage{booktabs}
\usepackage{array}
\newcolumntype{L}[1]{>{\raggedright\arraybackslash}p{#1}}
\usepackage{graphicx}
\usepackage{multirow}
\usepackage{xcolor}
\usepackage{enumitem}
\usepackage{wrapfig}
\usepackage{caption}

\newcommand{\Eq}[1]{Eq.~(\ref{eq:#1})}

\def\ie{\textit{i.e.}}

\title{Taming VLAs under Robot Execution Errors: \\
Self-Compensation and Stress Testing}

\author{\textbf{Sohyun Lee}$^{1*}$ \quad \textbf{Yoonjae Baek}$^{1*}$ \quad \textbf{Jaesang Won}$^{1}$ \quad \textbf{Jinnyeong Kim}$^{1}$ \\
\textbf{Hyunwoo Kang}$^{1}$ \quad \textbf{Seung-Hwan Baek}$^{1}$ \quad \textbf{Ivan Laptev}$^{2}$ \quad \textbf{Suha Kwak}$^{1}$ \\[3pt]
{\normalfont $^{1}$POSTECH \quad \normalfont $^{2}$MBZUAI} 
}

\iclrfinalcopy
\begin{document}

\maketitle
\lhead{Preprint.}
\renewcommand{\thefootnote}{\fnsymbol{footnote}}
\footnotetext[1]{Equal contribution.}
\renewcommand{\thefootnote}{\arabic{footnote}}

\begin{abstract}
Vision-language-action (VLA) policies often fail when a robot's executed motion deviates from their commanded action. Such execution errors arise from the robot's mechanics and operating conditions, such as wear and payload changes. We propose self-compensating VLA, a deployment-time adaptation method that enables a VLA policy to pre-compensate for the robot's execution errors when generating commands. Without task rewards or labels, it updates the policy online using the residual between the action commanded by a VLA and the motion executed by the robot. To stress-test VLA robustness across execution conditions that are impractical to cover with physical robots alone, we introduce RoboStress, a controlled simulation benchmark. It combines established joint-level models of friction, backlash, compliance, and gravity-compensation error into seven deployment scenarios whose execution errors depend on the robot's state and motion history. On RoboStress, self-compensating VLA achieves higher average task success than both the base policies and methods that build in robustness during training. On two physical robot arms with different usage histories, it raises the average task success rate by more than 30 percentage points on each arm, and the gains extend to objects not seen in the task demonstrations.

\end{abstract}

\section{Introduction}
\label{sec:intro}

Vision-language-action (VLA) models~\citep{zitkovich2023rt, kim2024openvla,black2024pi_0,black2025pi_,pertsch2025fast,bjorck2025gr00t,cheang2024gr,wen2025tinyvla} enable general-purpose robot manipulation, mapping camera observations and language instructions directly to robot actions across diverse tasks~\citep{brohan2023rt, octo_2023}.
As these models move toward practical deployment,
their robustness has gained increasing attention~\citep{fei2025libero, liu2025eva, hancock2025run, zhang2025robustvla, guo2025robustness}.
Existing studies~\citep{fei2025libero, liu2025eva, zhang2025gevrm} have primarily examined robustness to visual corruptions, distractors, and paraphrased instructions.
Recent work has extended this analysis to action perturbations by injecting synthetic noise into a policy's output commands~\citep{zhang2025robustvla, guo2025robustness}.

During physical deployment, the motion executed by a robot often differs from a VLA's action commands even without artificially injected perturbations. 
These execution errors stem from the robot's mechanics and operating conditions, such as wear, heavy payloads, and temperature changes~\citep{bittencourt2013modeling, gaz2017payload}.
Unlike the synthetic perturbations added to action commands in the previous work, these errors arise as the robot executes the commands.
Although such execution errors frequently occur in physical deployments, the robustness of VLA models to them remains largely underexplored.

\begin{figure}[t]
\centering
\includegraphics[width=1.0\linewidth]{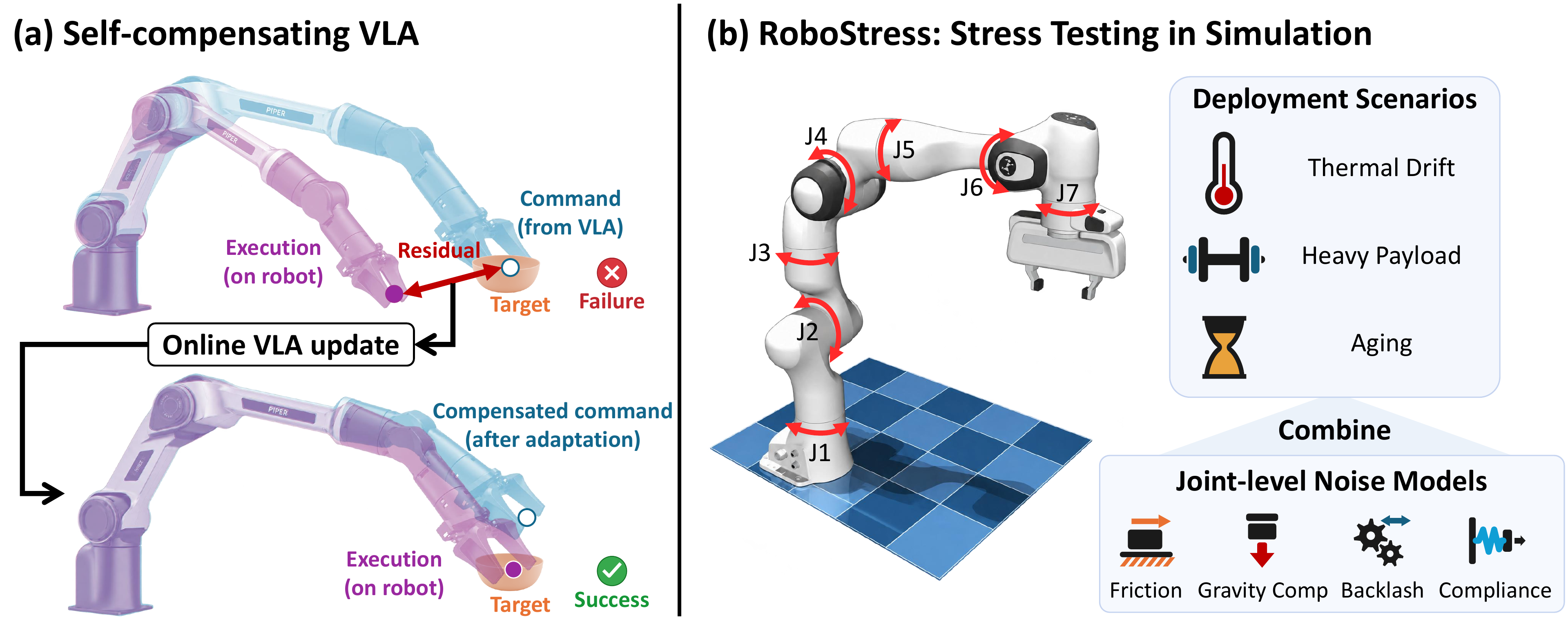}
\vspace{-4mm}
\caption{Overview of our study. (a) Self-compensating VLA updates the policy online using command-execution residuals and generates subsequent commands that pre-compensate for execution errors.
(b) The RoboStress benchmark combines four established joint-level models into controlled deployment scenarios in simulation to stress-test VLA robustness.
} 
\label{fig:teaser}
\end{figure}

Motivated by this, we study the robustness of VLA models to execution errors, as illustrated in Figure~\ref{fig:teaser}(a).
These errors vary across robots and operating conditions, calling for policy adaptation to the deployed robot.
Our key insight is that the residual between the action commanded by a VLA and the motion executed by the robot, measured through proprioception, can serve as
a self-supervisory signal for policy adaptation during deployment.
Based on this idea, we propose a self-compensation method for VLAs, dubbed \emph{self-compensating VLA}.
Our method adapts the action policy online using this residual, without task rewards, labels, or knowledge of the error sources.
The updated policy then generates subsequent commands that pre-compensate for observed execution errors while leaving the robot's low-level controller unchanged.

Ideally, evaluating VLA robustness to execution errors requires a fleet of robots with diverse mechanical states, usage histories, and operating conditions.
However, assembling such a fleet at scale is impractical, and varying operating conditions on physical robots is costly, with extreme settings posing a risk of hardware damage.
A natural alternative is to reproduce and systematically vary these conditions in simulation.
Existing simulated evaluations~\citep{zhang2025robustvla,guo2025robustness}, however, rely on synthetic action perturbations that differ from execution errors encountered during physical deployments in two respects.
First, real execution errors vary with both the robot's current state and its recent motion history, while the synthetic perturbations do not explicitly model such dependencies.
Second, errors at individual joints affect end-effector motion differently, whereas the synthetic perturbations applied directly to end-effector commands do not capture the distinct contribution of each joint.

To address these limitations, we introduce RoboStress, a controlled simulation benchmark for stress-testing VLA robustness across diverse execution conditions, as shown in Figure~\ref{fig:teaser}(b).
To be specific, RoboStress uses established joint-level models of friction~\citep{de1995new, madsen2020comprehensive}, backlash~\citep{tao1993adaptive}, compliance~\citep{spong1987modeling}, and gravity-compensation error~\citep{ma1996identifying, gaz2019dynamic}.
Each model is applied to individual joints at the corresponding stage of the control pipeline.
We combine these effects into seven deployment scenarios whose execution errors depend on the robot's state and recent motion history.

Extensive experiments on RoboStress and two physical robots validate the effectiveness of self-compensating VLA.
On RoboStress, it outperforms two base policies, $\pi_0$~\citep{black2024pi_0} and $\pi_{0.5}$~\citep{black2025pi_}, and methods that embed robustness during training~\citep{guo2025robustness,tobin2017domain} across all seven deployment scenarios.
Beyond simulation, we evaluate self-compensating VLA on two physical Piper arms with different usage histories.
For both base policies, it raises the average task success rate by more than 30 percentage points on both arms.
It also improves task success with objects not seen in the task demonstrations.
Our benchmark and method lay a foundation for future research on VLA robustness to robot execution errors.

\section{Related Work}
\label{sec:related}

\paragraph{VLA Robustness.}
Previous work evaluates VLAs under input perturbations such as viewpoint, lighting, or instruction changes~\citep{fei2025libero,liu2025eva}.
Other methods improve robustness through visual input interventions~\citep{hancock2025run} or video-based planning and state representation alignment~\citep{zhang2025gevrm}.
Recent work also studies robustness to synthetic perturbations applied to output end-effector commands~\citep{zhang2025robustvla,guo2025robustness}.
Our work instead addresses execution errors arising from robot mechanics and operating conditions.

\paragraph{Robot Dynamics Modeling.}
The joint-level models used in RoboStress are well established in robot dynamics: Coulomb-viscous and Stribeck-style friction laws~\citep{de1995new, lampaert2003generalized}, hysteretic deadband models of backlash~\citep{tao1993adaptive}, elastic-joint models of compliance~\citep{spong1987modeling, hardeman2008modelling}, and gravity-compensation error from inertial-parameter and payload mismatch~\citep{ma1996identifying, gaz2019dynamic}.
Joint friction varies with temperature and wear~\citep{bittencourt2013modeling,raviola2021effects}, and related dynamics models have been experimentally evaluated on the Franka Panda~\citep{gaz2019dynamic} and UR5e~\citep{madsen2020comprehensive}.
Whereas these models were developed for identification and control, RoboStress uses them to generate execution errors for evaluating VLA robustness.

\paragraph{Deployment-Time Adaptation.}
Earlier work adapts reinforcement learning policies during deployment through auxiliary self-supervised learning objectives~\citep{hansen2020self} or a pretrained adaptation module that infers deployment conditions from recent states and actions~\citep{kumar2021rma}.
EVOLVE-VLA~\citep{bai2025evolvevla} and TT-VLA~\citep{liu2026flyttvla} adapt VLAs at test time using task-progress-based feedback.
Our method adapts the action policy using the command-execution residual measured through proprioception, directly targeting the deployed robot's execution errors without task rewards.

\paragraph{Robot Control.}
Model-based controllers use robot dynamics and feedback to regulate motion and force in operational space~\citep{khatib1987unified}, while disturbance-observer-based methods estimate disturbances from feedback and compensate for them in the control input~\citep{chen2015disturbance}.
Adaptive control handles parametric uncertainty through online parameter estimation~\citep{slotine1987adaptive}.
Learning-based control learns compensation from feedback-controller outputs~\citep{gomi1993neural} or adds a reward-trained residual to a nominal controller~\citep{johannink2019residual}.
Self-compensating VLA compensates at the policy level by adapting the commands sent to the existing controller, without modifying that controller or requiring an explicit dynamics model.

\section{Self-compensating VLA}
\label{sec:method}

\begin{figure}[t]
\centering
 \includegraphics[width=1.0\linewidth]{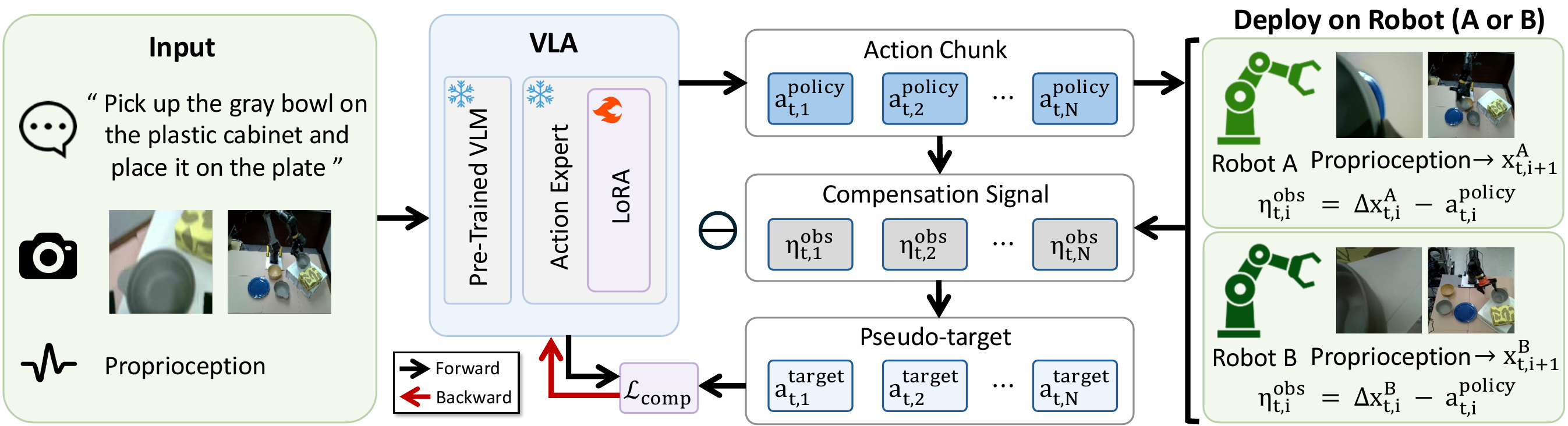}
 \vspace{-4mm}
\caption{
Overview of self-compensating VLA.
A frozen VLA with trainable LoRA adapters predicts an action chunk, which the robot executes.
The residual $\eta^\textrm{obs}$ between the commanded and executed motion forms a pseudo-target $a_t^{\mathrm{target}}$, which the compensation objective $\mathcal{L}_\textrm{comp}$ uses to update only the LoRA adapters online during deployment, leaving the base policy frozen.
}
\label{fig:method}
\end{figure}

Self-compensating VLA adapts a pretrained VLA to individual robots during deployment to improve its robustness against execution errors.
As illustrated in Figure~\ref{fig:method}, it first obtains a compensation signal from the residual between commanded and executed motion (\S\ref{sec:signal}), and then uses the signal to update the policy online so that subsequent commands pre-compensate for execution errors (\S\ref{sec:absorb}).

\subsection{Compensation Signal}
\label{sec:signal}
We consider a policy that outputs an action chunk, where each action $a_{t,i}^{\text{policy}}$ is a delta pose, \ie, the commanded change in end-effector pose at step $i$ of chunk $t$.
At each executed step, we measure the residual by subtracting the commanded delta pose from the executed one:
\begin{equation}
\eta_{t,i}^{\text{obs}} = \Delta x_{t,i} - a_{t,i}^{\text{policy}},
\label{eq:eta-obs}
\end{equation}
where $x_{t,i}$ is the end-effector pose at step $i$ of chunk $t$, measured through proprioception, and $\Delta x_{t,i}$ is the achieved change from $x_{t,i}$ to $x_{t,i+1}$.
The policy commands a normalized displacement $a_{t,i}^{\text{policy}}$, whereas proprioception measures the executed displacements in physical units.
We thus compute the rotational displacement using the SO(3) logarithm map~\citep{murray2017mathematical} and divide the measured displacements by the scale factors used to convert commands to physical units for execution, obtaining $\Delta x_{t,i}$ on the same scale as $a_{t,i}^{\text{policy}}$.
Since the residual $\eta^{\text{obs}}$ is computed from issued commands and proprioceptive measurements, it provides a self-supervisory signal for policy adaptation.

\subsection{Deployment-time Policy Adaptation}
\label{sec:absorb}

Given the compensation signal, we adapt the policy so that its subsequent commands pre-compensate for the execution errors it reveals.
We construct a compensated pseudo-target for this adaptation by subtracting the observed residual from each executed command:
\begin{equation}
a_{t,i}^{\text{target}} = a_{t,i}^{\text{policy}} - \eta_{t,i}^{\text{obs}}.
\label{eq:corrected}
\end{equation}
Since the residual $\eta_{t,i}^{\text{obs}}$ is observed only after execution, we cannot use $a_{t,i}^{\text{target}}$ to correct the current chunk but only to update the policy for subsequent chunks.
To this end, we apply LoRA~\citep{hu2022lora} to the action expert of each base policy~\citep{black2024pi_0,black2025pi_}, updating only the LoRA parameters $\phi$ online while keeping the pretrained weights frozen.
Since the base policies are flow-matching models~\citep{lipman2022flow}, we adopt a flow-matching objective that regresses the predicted velocity toward the target velocity defined by the target action chunk:
\begin{equation}
\mathcal{L}_{\text{comp}}
=
\mathbb{E}_{t,u,\epsilon}\!\left[
\frac{1}{Kd}
\big\|v_\theta(s_t,a_t^u,u)
- (\epsilon-a_t^{\text{target}})\big\|_2^2
\right]
+
\lambda_{\text{anc}}\|\phi-\phi_0\|_2^2,
\label{eq:flow-matching}
\end{equation}
where $K$ is the chunk length, $d$ is the action dimension, $a_t^{\text{target}}$ is the target action chunk, and $v_\theta$ is the policy's velocity field.
The input $s_t$ comprises the visual observations, proprioceptive state, and language instruction used to generate the original action chunk.
The noisy target is $a_t^u = u\epsilon + (1-u)a_t^{\text{target}}$ at flow-matching time $u \in [0,1]$, with $\epsilon \sim \mathcal{N}(0,I)$ and target velocity $\epsilon-a_t^{\text{target}}$.
The anchor term, weighted by $\lambda_{\text{anc}}$, penalizes deviations of the LoRA parameters $\phi$ from their initialization $\phi_0$, limiting parameter drift during online adaptation.
We accumulate target chunks from executed action chunks in an online buffer, over which the expectation in Eq.~\eqref{eq:flow-matching} is taken, and update the LoRA parameters during deployment using robot execution feedback.

\section{The RoboStress Benchmark}
\label{sec:bench}

RoboStress is a controlled simulation benchmark for stress-testing VLA robustness to execution errors.
It models four joint-level noise components (\S\ref{sec:components}), applies each at the corresponding stage of the control pipeline (\S\ref{sec:hybrid}), and combines them into seven deployment scenarios (\S\ref{sec:scenarios}).

\subsection{Joint-level Noise Models}
\label{sec:components}

We consider four noise components that cause execution errors: friction, gravity-compensation error, backlash, and compliance.
For each, we adopt a joint-level model from the robot dynamics literature~\citep{de1995new, madsen2020comprehensive, tao1993adaptive, spong1987modeling, gaz2019dynamic} and represent its effect on joint $j$ as a noise term $\eta_j$.

\noindent\textbf{Stribeck friction.}\label{sec:stribeck}
Joint friction depends nonlinearly on velocity and is elevated near zero speed~\citep{de1995new, lampaert2003generalized}.
We adopt the Coulomb-Stribeck law with a viscous term~\citep{de1995new}:
\begin{equation}
\eta^{\text{fric}}_j = -\operatorname{sgn}(\dot{q}_j) \left[F_{c,j} + (F_{s,j} - F_{c,j})\, e^{-(\dot{q}_j / v_{s,j})^2}\right] - \sigma_{v,j} \dot{q}_j,
\label{eq:stribeck}
\end{equation}
where $\dot{q}_j$ is the joint velocity, \ie, the time derivative of its angle $q_j$, $\operatorname{sgn}(\cdot)$ is the sign function, $F_{s,j}$ and $F_{c,j}$ are its static and kinetic friction torques,
$v_{s,j}$ the Stribeck velocity, and $\sigma_{v,j}$ the viscous coefficient.
We add the resulting torque $\eta^{\text{fric}}_j$ to the joint torque computed by the controller.
Since $F_{s,j}$ dominates at low speed $|\dot{q}_j|$, small commands struggle against static friction, degrading low-speed tracking.

\noindent\textbf{Gravity-compensation error.}\label{sec:grav}
The controller compensates for gravity using estimated link masses, and the mismatch with the true link masses, together with any unknown payload, leaves a configuration-dependent residual torque on each joint~\citep{gaz2019dynamic}.
We inject this residual as a fraction of the gravitational torque computed by the controller, denoted by $\eta^{\text{grav}}_j$ on joint $j$: $\eta^{\text{grav}}_j = \beta_j \, g_j(\mathbf{q}),$
where
$g_j(\mathbf{q})$ is the gravitational torque computed by the controller at the joint angles $\mathbf{q} = (q_1, \dots, q_n)$ of the $n$ joints, and $\beta_j \geq 0$ is the residual fraction controlling the strength of the error.
$\eta^{\text{grav}}_j$ depends only on the current joint angles $\mathbf{q}$, not on motion history.

\noindent\textbf{Backlash.}\label{sec:backlash}
Gear clearance, which widens with wear~\citep{bittencourt2013modeling}, creates a deadband on direction reversal in which the motor rotates but the
link stays still.
We adopt the Tao-Kokotovi\'{c} deadband model~\citep{tao1993adaptive}, which gives the link angle $q^l_j$ of joint $j$ as a function of its motor angle $q^m_j$:
\begin{equation}
q^l_j(t) = \begin{cases}
q^m_j(t) - B_j & \text{if } q^m_j(t) - q^l_j(t{-}1) > B_j,\\
q^m_j(t) + B_j & \text{if } q^m_j(t) - q^l_j(t{-}1) < -B_j,\\
q^l_j(t{-}1)   & \text{otherwise,}
\end{cases}
\label{eq:backlash}
\end{equation}
where $t$ indexes the time step and $B_j$ is the half-width of the deadband at joint $j$.
We inject backlash by replacing $q^m_j$ with the deflected angle $q^l_j$ from Eq.~\eqref{eq:backlash} as the joint position that enters the simulator's integration step (\S\ref{sec:hybrid}).
The injected noise $\eta^{\text{back}}_j$ represents the angular deviation of the link from the motor due to backlash: $\eta^{\text{back}}_j = q^l_j(t) - q^m_j(t),$
which is bounded by the deadband half-width, $|\eta^{\text{back}}_j| \le B_j$.
$\eta^{\text{back}}_j$ is history-dependent because it depends on the previous link angle $q^l_j(t{-}1)$ as well as the current motor angle $q^m_j(t)$.

\noindent\textbf{Dynamic compliance.}\label{sec:compliance}
A robot joint is not perfectly rigid but compliant, which causes the link to deflect under load, deviating from the commanded angle when the joint torque changes~\citep{spong1987modeling, hardeman2008modelling}.
We take such angular deflection as a type of noise.
The deflection on joint $j$, denoted $\eta^{\text{comp}}_j$, is formulated as a spring-mass-damper response~\citep{spong1987modeling} to the load torque:
\begin{equation}
M_j^{\text{eff}}\,\ddot{\eta}^{\text{comp}}_j + D_j\,\dot{\eta}^{\text{comp}}_j + K_j\,\eta^{\text{comp}}_j = -\tau_j^{\text{load}},
\label{eq:compliance}
\end{equation}
where $\dot{\eta}^{\text{comp}}_j$ and $\ddot{\eta}^{\text{comp}}_j$ are its angular velocity and acceleration, $\tau_j^{\text{load}}$ is the torque applied at the joint, and $M_j^{\text{eff}}$, $D_j$, and $K_j$ are the effective inertia, damping, and stiffness of the joint, respectively.
The negative sign reflects that the load compresses the transmission, so the link lags behind the motor.
The second-order dynamics can produce transient overshoot and oscillation under abrupt load changes.
The deflection does not have a fixed value at each step but evolves over time, and we compute it by integrating Eq.~\eqref{eq:compliance} online.
$\eta^{\text{comp}}_j$ is history-dependent because the deflection reflects the joint's load history, not just the current load.

\begin{figure}[t]
\centering
 \includegraphics[width=1.0\linewidth]{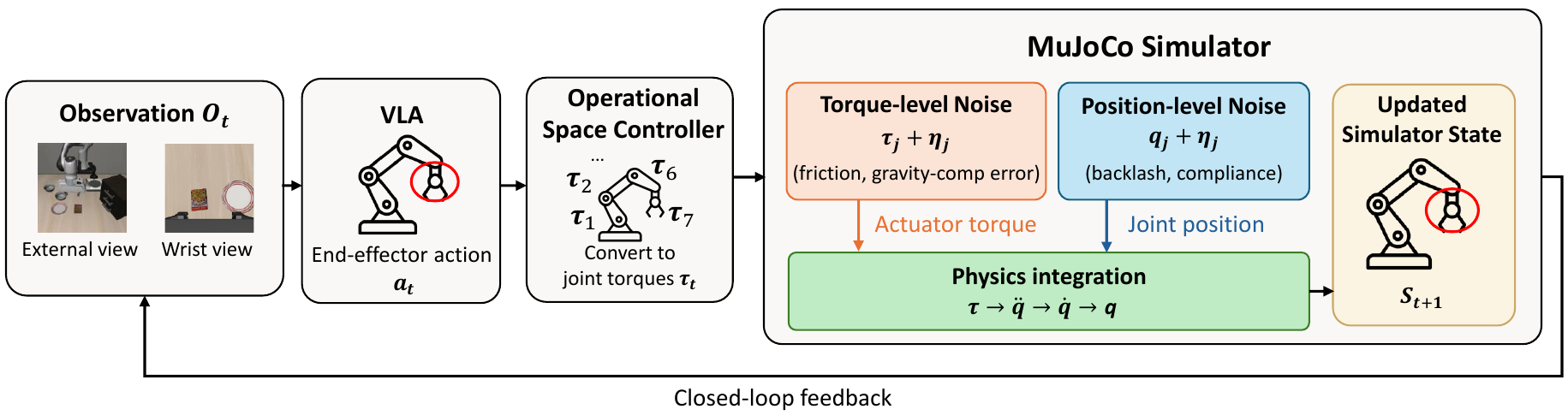}
 \vspace{-4mm}
\caption{Overview of the joint-level noise injection. The operational space controller~\citep{khatib1987unified} converts end-effector commands into joint torques. 
Torque-level noise (friction, gravity-compensation error) is added to joint torques, and position-level noise (backlash, compliance) to joint positions before physics integration.
The next observation is rendered from the updated state.}
\label{fig:hybrid}
\end{figure}

\subsection{Joint-level Noise Injection}
\label{sec:hybrid}
As shown in Figure~\ref{fig:hybrid}, we inject each noise term $\eta_j$ where it physically arises.
The operational space controller~\citep{khatib1987unified} converts the Cartesian action into joint torques $\tau_j$.
Friction and gravity-compensation error are added to this torque, and $\tau^{\text{load}}_j = \tau_j + \eta^{\text{fric}}_j + \eta^{\text{grav}}_j$ is applied to the actuator.
Backlash and compliance are added to the joint position that enters the simulator's integration step, $q^m_j + \eta^{\text{back}}_j + \eta^{\text{comp}}_j$, with the compliance driven by $\tau^{\text{load}}_j$ through Eq.~\eqref{eq:compliance}.
The forward dynamics then propagates the perturbed joint torques and positions through the arm, so a perturbation at one joint affects other joints and produces different end-effector errors depending on the arm's pose.

\subsection{Deployment Scenarios}
\label{sec:scenarios}
\begin{table}[t]
\centering
\caption{Deployment scenarios in RoboStress.
Checked noise components have increased severity, while unmarked components retain their nominal settings.
}
\vspace{-2mm}
\label{tab:scenarios}
\small
\setlength{\tabcolsep}{3pt}
\renewcommand{\arraystretch}{1.0}
\begin{tabular*}{\linewidth}{@{\extracolsep{\fill}}llccccll@{}}
\toprule
Scenario & Physical cause
& Fric. & Grav. & Back. & Comp.
& Severity over episode & Joints \\
\midrule
Heavy Payload
& grasped load
& & $\checkmark$ & & $\checkmark$
& While grasping & all \\

Thermal Drift-Stribeck
& temperature change
& $\checkmark$ & & &
& Linearly increasing & all \\

Thermal Drift-Backlash
& temperature change
& & & $\checkmark$ &
& Linearly increasing & all \\

Aged Transmission
& gear wear
& $\checkmark$ & & $\checkmark$ &
& Fixed & all \\

Aged Joint-Uniform
& gear + bearing wear
& $\checkmark$ & & $\checkmark$ & $\checkmark$
& Fixed & all \\

Aged Joint-Shoulder
& gear + bearing wear
& $\checkmark$ & & $\checkmark$ & $\checkmark$
& Fixed & J1--J3 \\

Aged Joint-Elbow
& gear + bearing wear
& $\checkmark$ & & $\checkmark$ & $\checkmark$
& Fixed & J4 \\
\bottomrule
\end{tabular*}
\end{table}

A deployed robot rarely exhibits a single noise component of \S\ref{sec:components} in isolation.
A robot that is worn, warming up, or carrying a heavy payload is affected by several of these noise components simultaneously, with magnitudes that differ across joints and may change within an episode.
RoboStress therefore combines the four noise components and varies their severity across the seven deployment scenarios of Table~\ref{tab:scenarios}, which together probe different aspects of a policy's robustness.

\noindent\textbf{Severity factors.\label{sec:wear}}
Each scenario controls the strength of the four noise components through per-joint severity factors that scale their model parameters in \S\ref{sec:components}:
\begin{equation}
(F_{s,j},F_{c,j},\sigma_{v,j})
=w^f_j(F_{s,j,0},F_{c,j,0},\sigma_{v,j,0}),
\; \beta_j=w^g_j\beta_{j,0},
\; B_j=w^b_jB_{j,0},
\; K_j=K_{j,0}/w^c_j,
\label{eq:wear}
\end{equation}
where the subscript $0$ denotes the parameter value before severity scaling, and $w^{f}_j$, $w^{g}_j$, $w^{b}_j$, and $w^{c}_j \ge 1$ are the severity factors for friction, gravity-compensation error, backlash, and compliance at joint $j$, respectively.
The nominal setting $w=1$ leaves the model parameters unchanged.
As shown in Table~\ref{tab:scenarios}, each deployment scenario specifies which components have increased severity, which joints are affected, and whether the factors remain fixed or increase linearly over an episode.
The factor values for all scenarios are provided in Appendix~\ref{sec:supp_composite}.

\noindent\textbf{Heavy Payload.}
\emph{Heavy Payload} models a grasped payload, causing gravity-compensation error and larger joint deflections.
We emulate these effects by increasing $w^{g}_j$ and $w^{c}_j$ on every joint, while the gripper holds an object, keeping friction and backlash nominal.
This scenario therefore tests robustness to execution errors that depend on the arm's pose and load.

\noindent\textbf{Thermal Drift.}
\emph{Thermal Drift} models execution errors that change with a robot's temperature~\citep{bittencourt2013modeling}.
\emph{Thermal Drift-Stribeck} and \emph{Thermal Drift-Backlash} linearly increase $w^{f}_j$ and $w^{b}_j$, respectively, over the episode.
The two variants separately model gradual changes in a velocity-dependent torque component and a history-dependent position component.
They test policy adaptation to errors that change within an episode.

\noindent\textbf{Aged Transmission and Aged Joint.}
Mechanical wear degrades gears and bearings, increasing friction, backlash, and compliance unevenly across joints~\citep{bittencourt2013modeling}.
\emph{Aged Transmission} models gear wear, where worn teeth increase backlash and degraded lubrication raises friction.
We therefore increase $w^{f}_j$ and $w^{b}_j$ on every joint while keeping compliance nominal.
\emph{Aged Joint} additionally models bearing wear by increasing $w^{c}_j$, with variants that apply the same severity factors to different sets of joints.
\emph{Aged Joint-Uniform} applies these factors to all seven joints, \emph{Aged Joint-Shoulder} to joints J1-J3 near the base, and \emph{Aged Joint-Elbow} to joint J4 at the elbow.
These three variants test how policy robustness depends on which joints are affected by wear.

\section{Experiments}
\label{sec:exp}

\subsection{Experimental Setting}
\label{sec:setting}
\noindent\textbf{Implementation Details.}
We adopt $\pi_0$~\citep{black2024pi_0} and $\pi_{0.5}$~\citep{black2025pi_} as base policies for both the RoboStress and real-world experiments; unless otherwise stated, analyses use $\pi_0$.
For self-compensating VLA, we attach a rank-4 LoRA~\citep{hu2022lora} to each base policy's action expert and optimize it with Adam~\citep{kingma2014adam} at a learning rate of $10^{-6}$ and an anchor weight of $\lambda_{\text{anc}}=10^{-4}$.
For real-world experiments, we fine-tune each base policy and RobustVLA on teleoperated demonstrations collected on each Piper arm.
On the Piper arms, the policies output absolute joint positions, so we compute residuals and compensated targets in joint space.
More implementation details are given in Appendix~\ref{sec:supp_impl}.

\noindent\textbf{Evaluation Protocol.}
We report the success rate averaged over tasks.
On RoboStress, which spans all four task suites of LIBERO~\citep{liu2023libero} with 10 tasks each, we evaluate the seven deployment scenarios, using 20 episodes per task in each setting; parameter settings of noise components are detailed in Appendix~\ref{sec:supp_benchmark}.
For real-world experiments, we evaluate on two AgileX Piper arms with naturally occurring execution errors (\S\ref{sec:real}), using 25 episodes per task.

\noindent\textbf{Competitors.}
We compare against two methods applied to improve VLA robustness, both of which operate at training time.
Domain randomization (DR)~\citep{tobin2017domain}, applied here to the action space, fine-tunes $\pi_0$ and $\pi_{0.5}$ under Gaussian noise ($\sigma\sim\mathcal{U}[0,0.05]$) on the end-effector command, while RobustVLA~\citep{guo2025robustness} adversarially trains each base policy within an $\ell_\infty$ $\varepsilon$-ball ($\varepsilon=0.03$, PGD of~\citet{madry2017towards}).

\begin{table}[t]
\centering
\caption{Success rates (\%) on RoboStress deployment scenarios, averaged over LIBERO tasks.
Averages exclude Clean.
Best results within each backbone are shown in bold for each row.}
\vspace{-3mm}
\label{tab:realistic}
\small
\setlength{\tabcolsep}{3pt}
\renewcommand{\arraystretch}{1.0}
\begin{tabular*}{\linewidth}{@{\extracolsep{\fill}}lcccccccc@{}}
\toprule
& \multicolumn{4}{c}{$\pi_{0.5}$}
& \multicolumn{4}{c}{$\pi_0$} \\
\cmidrule(lr){2-5}\cmidrule(lr){6-9}
Scenario
& Base & DR & RobustVLA & Ours
& Base & DR & RobustVLA & Ours \\
\midrule
Clean
& 97.0 & \textbf{97.1} & 96.6 & 95.8
& 91.1 & 90.5 & 92.0 & \textbf{92.6} \\
\midrule
Heavy Payload
& 44.1 & 42.1 & 38.9 & \textbf{56.8}
& 42.3 & 32.0 & 30.9 & \textbf{42.8} \\
Thermal Drift-Stribeck
& 54.1 & 54.6 & 55.9 & \textbf{60.5}
& 46.1 & 46.3 & 46.0 & \textbf{50.6} \\
Thermal Drift-Backlash
& 83.4 & 84.5 & 83.1 & \textbf{84.8}
& 75.5 & 77.3 & 77.3 & \textbf{78.8} \\
Aged Transmission
& 33.2 & 39.7 & 40.2 & \textbf{41.8}
& 26.8 & 22.9 & 23.8 & \textbf{28.9} \\
Aged Joint-Uniform
& 29.5 & 36.0 & 33.8 & \textbf{38.6}
& 23.0 & 22.4 & 20.0 & \textbf{24.0} \\
Aged Joint-Shoulder
& 43.9 & 46.4 & 43.0 & \textbf{51.2}
& 34.5 & 32.9 & 31.5 & \textbf{39.2} \\
Aged Joint-Elbow
& 55.6 & 54.0 & 53.9 & \textbf{62.3}
& 43.9 & 44.5 & 42.2 & \textbf{44.9} \\
\midrule
\textbf{Avg.}
& 49.1 & 51.0 & 49.8 & \textbf{56.6}
& 41.7 & 39.8 & 38.8 & \textbf{44.2} \\
\bottomrule
\end{tabular*}
\end{table}

\subsection{Experiments on RoboStress}
\label{sec:main-results}
\noindent\textbf{Results on deployment scenarios.}
Table~\ref{tab:realistic} shows that self-compensating VLA outperforms the base policy, DR, and RobustVLA in all seven scenarios with both backbones.
It raises average success from 49.1\% to 56.6\% with $\pi_{0.5}$ and from 41.7\% to 44.2\% with $\pi_0$.
In contrast, DR and RobustVLA yield inconsistent gains and fall below the base policy on average with $\pi_0$.

\begin{figure}[t]
\centering
\begin{minipage}[c]{0.66\linewidth}
    \centering
    \includegraphics[width=\linewidth]{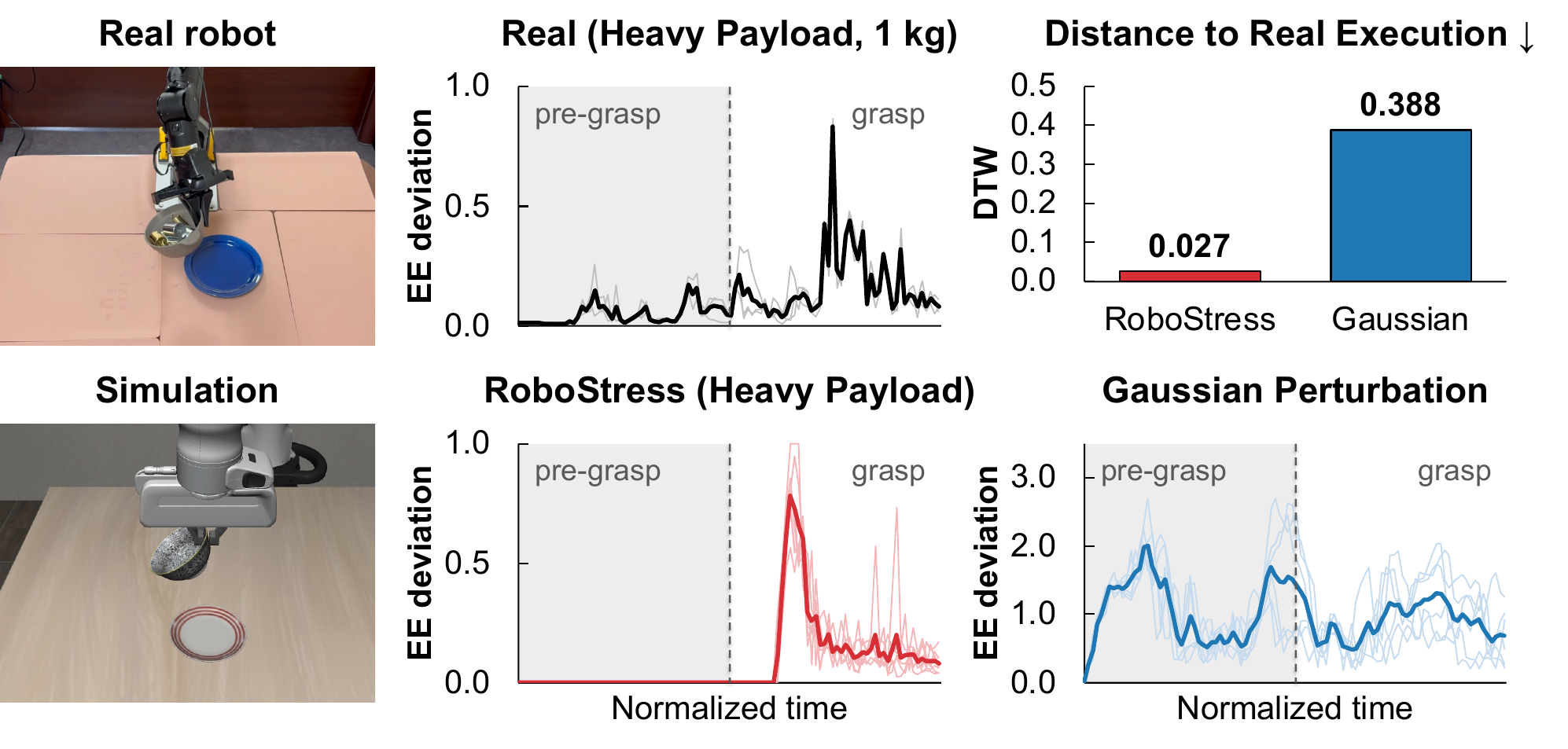}
    \vspace{-6mm}
    \caption{Real-world payload execution deviations compared with RoboStress and Gaussian action noise.}
    \label{fig:fidelity}

\end{minipage}
\hfill
\begin{minipage}[c]{0.33\linewidth}
    \centering
    \captionof{table}{Noise analysis.}
    \vspace{-3mm}
    \footnotesize
    \renewcommand{\arraystretch}{0.9}
    \resizebox{0.87\linewidth}{!}{
    \begin{tabular}{l cc}
    \toprule
    Noise components & Base & Ours \\
    \midrule
    Stribeck   & 36.8 & \textbf{37.8} \\
    Backlash   & 68.0 & \textbf{72.1} \\
    Compliance & 74.1 & \textbf{74.2} \\
    Gravity    & 21.1 & \textbf{22.0} \\
    \midrule
    \textbf{Avg.} & 50.0 & \textbf{51.5} \\
    \bottomrule
    \end{tabular}}
    \label{tab:unit}
    \vspace{-1mm}
    \setlength{\tabcolsep}{4pt}
    \captionof{table}{Method ablation.}
    \label{tab:no-residual}
    \renewcommand{\arraystretch}{0.98}
    \resizebox{\linewidth}{!}{%
    \begin{tabular}{l cccc}
        \toprule
        & Base & Direct & w/o $\eta^{\text{obs}}$ & Ours \\
        \midrule
        Success (\%) & 41.7 & 23.6 & 40.8 & \textbf{44.2} \\
        \bottomrule
    \end{tabular}
    }
    \vspace{-1mm}
    \setlength{\tabcolsep}{4pt}
    \captionof{table}{
        Effect of DOB.
    }
    \renewcommand{\arraystretch}{0.98}
    \resizebox{\linewidth}{!}{%
    \begin{tabular}{l cccc}
        \toprule
        & Base & DOB & Ours & DOB+Ours \\
        \midrule
        Success (\%) & 48 & 56 & \textbf{84} & 80 \\
        \bottomrule
    \end{tabular}
    }
    \label{tab:feedback_comparison}
\end{minipage}
\end{figure}

\noindent\textbf{Comparing RoboStress Heavy Payload with real payload execution.}
We replay a pick-and-place demonstration three times on a Piper arm carrying a bowl with 1\,kg of weights, and adjust the Heavy Payload parameters to match the resulting end-effector deviations from an unloaded replay.
We then replay the task from ten initial layouts in simulation under the adjusted model and under Gaussian action noise at the scale used in RobustVLA's evaluation~\citep{guo2025robustness}, measuring deviations relative to the corresponding unperturbed replays.
As shown in Figure~\ref{fig:fidelity}, the adjusted RoboStress model captures the post-grasp rise in deviation and yields a lower dynamic time warping (DTW) distance to the real mean profile than the Gaussian reference.

\begin{table}[t]
\centering
\caption{
Real-world success rates (\%) on two Piper arms.
\textbf{Tasks 1--3} involve picking up the gray bowl from next to the plate, on the plastic cabinet, and on the gift box, respectively, and placing it on the plate.
\textbf{Task 4} involves opening the cabinet's top drawer.
Best results are shown in bold.
}
\vspace{-2mm}
\setlength{\tabcolsep}{3.5pt}
\renewcommand{\arraystretch}{1.0}
\resizebox{\linewidth}{!}{%
\begin{tabular}{l ccc ccc ccc ccc}
    \toprule
    & \multicolumn{6}{c}{Robot A}
    & \multicolumn{6}{c}{Robot B} \\
    \cmidrule(lr){2-7}
    \cmidrule(lr){8-13}
    & \multicolumn{3}{c}{$\pi_{0.5}$}
    & \multicolumn{3}{c}{$\pi_0$}
    & \multicolumn{3}{c}{$\pi_{0.5}$}
    & \multicolumn{3}{c}{$\pi_0$} \\
    \cmidrule(lr){2-4}
    \cmidrule(lr){5-7}
    \cmidrule(lr){8-10}
    \cmidrule(lr){11-13}
    Task
    & Base & RobustVLA & Ours
    & Base & RobustVLA & Ours
    & Base & RobustVLA & Ours
    & Base & RobustVLA & Ours \\
    \midrule
    1
    & 48 & 52 & \textbf{84}
    & 44 & 52 & \textbf{76}
    & 44 & 56 & \textbf{80}
    & 36 & 56 & \textbf{60} \\
    2
    & 36 & 60 & \textbf{76}
    & 40 & 28 & \textbf{80}
    & 36 & 48 & \textbf{72}
    & 32 & 52 & \textbf{84} \\
    3
    & 48 & 48 & \textbf{76}
    & 44 & 68 & \textbf{72}
    & 56 & 52 & \textbf{84}
    & 52 & 44 & \textbf{84} \\
    4
    & 40 & 48 & \textbf{68}
    & 32 & 40 & \textbf{60}
    & 36 & 44 & \textbf{60}
    & 32 & 44 & \textbf{56} \\
    \midrule
    \textbf{Avg.}
    & 43 & 52 & \textbf{76}
    & 40 & 47 & \textbf{72}
    & 43 & 50 & \textbf{74}
    & 38 & 49 & \textbf{71} \\
    \bottomrule
\end{tabular}%
}
\label{tab:real}
\end{table}

\begin{figure}[t]
\centering
\includegraphics[width=1.0\linewidth]{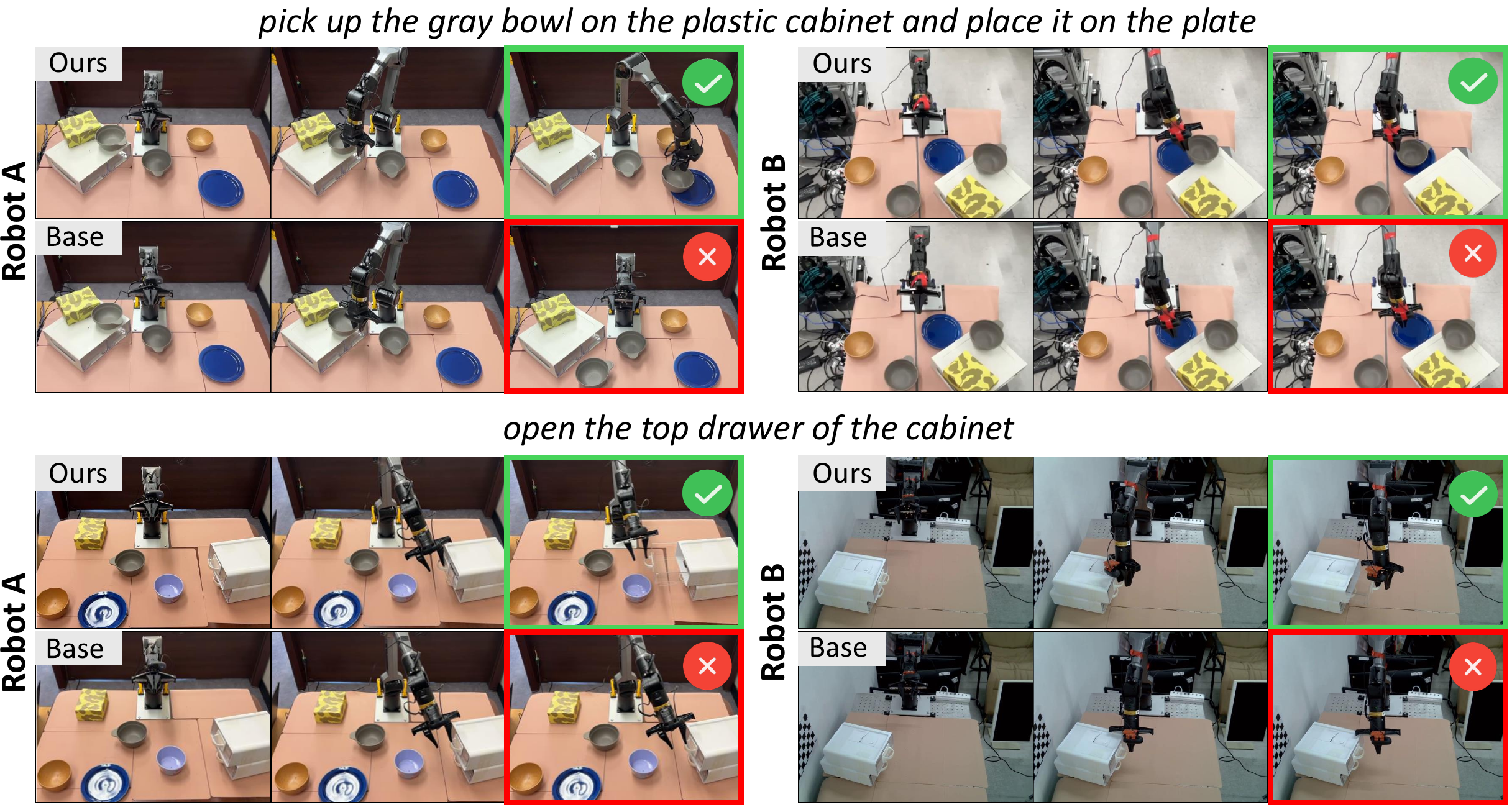}
\vspace{-2mm}
\caption{Rollouts of self-compensating VLA (Ours) and the base policy (Base) on Tasks 2 and 4, for Robot A and Robot B. Green and red borders mark success and failure.}
\label{fig:real}
\end{figure}

\noindent\textbf{Analysis on noise components.}
We evaluate each noise component in Table~\ref{tab:unit}.
Our method raises the average success rate of $\pi_0$ from 50.0\% to 51.5\%.
The smallest gain occurs under compliance, where the base policy already achieves its highest success rate.

\noindent\textbf{Effect of the residual and policy adaptation.}
We ablate components in our method on $\pi_0$ in Table~\ref{tab:no-residual}. Setting $\eta^{\text{obs}}{=}0$ retains online updates but uses the policy's commands as targets. This variant reaches 40.8\%, below the base policy and our method. Direct residual correction keeps the policy fixed and subtracts the mean residual of the previous chunk from the next chunk's commands, reaching only 23.6\%. Thus, the residual supplies the correction signal, while policy adaptation produces observation-conditioned compensation beyond the fixed chunk-level correction.

\subsection{Real-world Experiments}
\label{sec:real}
\noindent\textbf{Results on two real robot arms.}
We evaluate self-compensating VLA on a new AgileX Piper arm (Robot A) and another used for one year (Robot B).
The evaluation covers three bowl pick-and-place tasks and one drawer-opening task, as described in Table~\ref{tab:real}.
Even the new arm exhibits command-execution mismatch, with a mean normalized residual of 32.9\%, compared with 35.0\% on the older arm.
As shown in Table~\ref{tab:real}, self-compensating VLA outperforms the base policies and RobustVLA on every task with both $\pi_0$ and $\pi_{0.5}$ on both arms.
The representative rollouts in Figure~\ref{fig:real} show our method completing both tasks, whereas the base policy fails to grasp the bowl in Task 2 and to open the drawer in Task 4.
These results show that self-compensating VLA adapts to naturally occurring execution errors on two physical robots with different usage histories.

\begin{figure}[t]
\centering
\includegraphics[width=1.0\linewidth]{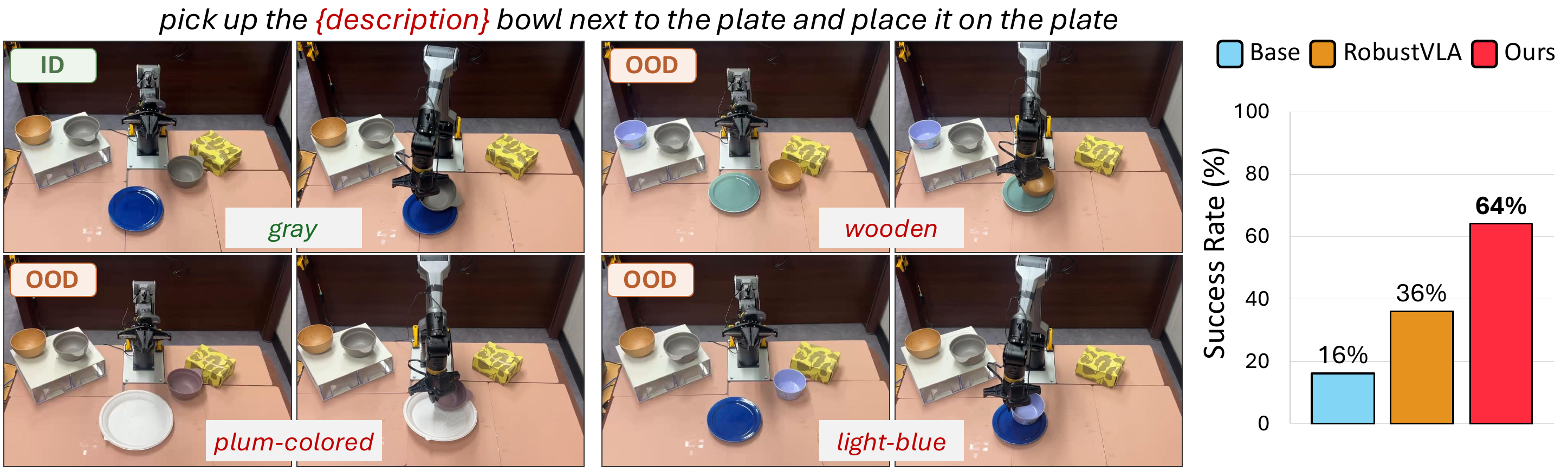}
\vspace{-2mm}
\caption{Generalization to unseen objects, evaluated on Robot A. Left: Our rollouts with the gray bowl (ID) and object variants (OOD). Right: Success rates on all OOD variants.}
\label{fig:ood}
\end{figure}

\noindent\textbf{Adaptation with external command correction.}
We evaluate whether our adaptation still helps under external command correction.
We add a disturbance observer (DOB)~\citep{chen2015disturbance} that uses joint feedback to estimate execution errors and correct commands before the Piper arm's built-in joint-position controller.
As shown in Table~\ref{tab:feedback_comparison}, on Task 1 of Robot A using $\pi_{0.5}$, DOB alone raises success from 48\% to 56\%, while adding our adaptation raises it further to 80\%, compared with 84\% for our adaptation alone.
These results indicate that our policy adaptation remains effective alongside external command correction and provides substantial gains beyond DOB alone.

\noindent\textbf{Generalization to objects unseen during fine-tuning.}
We evaluate Task 1 on Robot A with five object variants absent from the teleoperation fine-tuning data.
These variants change the appearance of the bowl or plate and include wooden, plum-colored, and light-blue bowls.
We update the instruction to describe each variant and run five episodes per variant, for 25 episodes in total.
As shown in Figure~\ref{fig:ood}, self-compensating VLA achieves $64\%$ success with $\pi_{0.5}$, compared with $16\%$ for the base policy and $36\%$ for RobustVLA, showing that its gains extend to unseen object variants.

\section{Conclusion}
\label{sec:conclusion}
We have studied VLA robustness to execution errors, a challenge for reliable robot deployment.
We have proposed self-compensating VLA, which adapts policies online using command-execution residuals to compensate for execution errors without task rewards or labels.
We have also introduced RoboStress, a simulation benchmark combining four joint-level error models into seven deployment scenarios.
Experiments on RoboStress and two physical Piper arms show improved task success over base policies and training-time robustness methods, with real-world gains extending to objects absent from the fine-tuning demonstrations.
These results demonstrate the effectiveness of deployment-time adaptation for improving VLA robustness to robot execution errors.
Future work includes incorporating motion estimates from external sensors to reduce reliance on proprioception.

\subsection*{AI use statement}
In this work, we used generative AI tools to provide feedback on experimental design, assist with the interpretation of results, and implement methods through code generation and debugging. We have not used generative AI tools to develop the core research ideas or derive the mathematical formulations.
We also used them to assist with translation, edit text and captions, and refine the manuscript structure.
We reviewed and revised AI-assisted text, checked suggested interpretations against experimental results, and verified AI-assisted code used in the experiments.
We take responsibility for the final content of this work, including text, claims or artifacts produced with the aid of generative AI.

\subsection*{Ethics statement}
This work aims to improve the reliability of robot manipulation under execution errors. RoboStress enables controlled stress testing in simulation, reducing the need to expose physical robots to potentially damaging conditions. Our real-world experiments evaluate manipulation tasks on two robot arms. The reported improvements in task success do not constitute safety guarantees; deployment in human environments requires additional safety evaluation and appropriate safeguards.

\subsection*{Reproducibility statement}
Section~3 describes the compensation signal, pseudo-target construction, and online adaptation objective. Section~4 and Appendix~A detail the RoboStress models, their integration into the simulation pipeline, and the parameters defining the deployment scenarios. Section~5.1 and Appendix~B document the evaluation protocol and implementation details, including the action and observation representations, LoRA configuration, online buffer, optimization settings, and computational setup. Appendix~D provides additional experimental analyses.
We will release the code for self-compensating VLA and RoboStress, including scenario configurations and evaluation scripts.

\bibliography{cvlab_kwak}
\bibliographystyle{iclr2027_conference}

\newpage
\appendix
\setcounter{table}{0}
\setcounter{figure}{0}
\setcounter{equation}{0}
\renewcommand{\thetable}{\Alph{section}\arabic{table}}
\renewcommand{\thefigure}{\Alph{section}\arabic{figure}}
\renewcommand{\theequation}{\Alph{section}\arabic{equation}}

\section{RoboStress Benchmark Details}
\label{sec:supp_benchmark}

\subsection{Noise Parameters}
\label{sec:supp_base_params}
Table~\ref{tab:base_params} lists the nominal values of the noise parameters defined in \S\ref{sec:components} for each joint $j$ of the 7-DoF arm.
The subscript $0$ marks a nominal value, before the severity factors of \Eq{wear} are applied.
The parameters include those of Stribeck friction ($F_{c,j,0}$, $F_{s,j,0}$, $v_{s,j}$, $\sigma_{v,j,0}$; \Eq{stribeck}), the backlash half-width ($B_{j,0}$; \Eq{backlash}), those of dynamic compliance ($M^{\text{eff}}_j$, $D_{j,0}$, $K_{j,0}$; \Eq{compliance}), and the gravity-compensation residual fraction ($\beta_{j,0}$; \Eq{gravity}).
Parameters without the subscript $0$, $v_{s,j}$ and $M^{\text{eff}}_j$, are not scaled by the severity factors.
Units are given in the table header, and $\beta_{j,0}$ is dimensionless.

The damping is set to $D_{j,0}=1.4\sqrt{K_{j,0}M_j^{\text{eff}}}$, yielding a damping ratio $\zeta=D_{j,0}/(2\sqrt{K_{j,0}M_j^{\text{eff}}})=0.70$ and a natural frequency $f_n=\frac{1}{2\pi}\sqrt{K_{j,0}/M_j^{\text{eff}}}$ of $14.7$\,Hz (J1--J4) and $22.5$\,Hz (J5--J7), corresponding to an underdamped second-order response.
The backlash half-width $B_{j,0}$ corresponds to approximately $0.05^{\circ}$ (J1--J4) and $0.075^{\circ}$ (J5--J7).
The residual fraction $\beta_{j,0}$ ranges from $5\%$ to $10\%$, with the largest values assigned to the shoulder pitch (J2) and elbow (J4).

\begin{table}[!ht]
\centering
\small
\setlength{\tabcolsep}{4pt}
\begin{tabular}{lccccccccc}
\toprule
& $F_{c,j,0}$ & $F_{s,j,0}$ & $v_{s,j}$ & $\sigma_{v,j,0}$ & $B_{j,0}$ & $M_j^{\text{eff}}$ & $D_{j,0}$ & $K_{j,0}$ & $\beta_{j,0}$ \\
Joint & (Nm) & (Nm) & (rad/s) & (Nm$\cdot$s/rad) & (rad) & (kg$\cdot$m$^2$) & (Nm$\cdot$s/rad) & (Nm/rad) & \\
\midrule
J1 & 2.175 & 3.480 & 0.15 & 0.348 & 0.00087 & 0.70 & 90.73 & 6000 & 0.05 \\
J2 & 2.175 & 3.480 & 0.15 & 0.348 & 0.00087 & 0.70 & 90.73 & 6000 & 0.10 \\
J3 & 2.175 & 3.480 & 0.15 & 0.348 & 0.00087 & 0.70 & 90.73 & 6000 & 0.06 \\
J4 & 2.175 & 3.480 & 0.15 & 0.348 & 0.00087 & 0.70 & 90.73 & 6000 & 0.10 \\
J5 & 0.480 & 0.720 & 0.10 & 0.048 & 0.00130 & 0.20 & 39.60 & 4000 & 0.08 \\
J6 & 0.480 & 0.720 & 0.10 & 0.048 & 0.00130 & 0.20 & 39.60 & 4000 & 0.08 \\
J7 & 0.480 & 0.720 & 0.10 & 0.048 & 0.00130 & 0.20 & 39.60 & 4000 & 0.05 \\
\bottomrule
\end{tabular}
\caption{Nominal noise parameters per joint $j$, defined in
\S\ref{sec:components}.}
\label{tab:base_params}
\end{table}

\subsection{Severity Factors}
\label{sec:supp_severity}

\Eq{wear} scales the noise parameters by per-joint severity factors $w^f_j$, $w^b_j$, $w^c_j$, $w^g_j \ge 1$.
Table~\ref{tab:severity} lists the full setting of each factor, namely $w^f = 2$, $w^b = 15$, $w^c = 5$, and $w^g = 4$.
For compliance, we also scale the damping as $D_j=D_{j,0}/\sqrt{w^c_j}$ alongside the stiffness scaling
$K_j=K_{j,0}/w^c_j$ of \Eq{wear} (with $M_j^{\mathrm{eff}}$ fixed), so that the damping ratio $\zeta=0.70$ is preserved as the stiffness drops. The payload factor $w^{g}_j$ scales the residual fraction $\beta_{j,0}$ and is used in Heavy Payload and in the gravity-compensation setting of Table~\ref{tab:isolated}.

\begin{table}[!ht]
\centering
\small
\setlength{\tabcolsep}{4pt}
\begin{tabular}{@{}l L{7.2cm} c@{}}
\toprule
Component & Applied to & Full setting \\
\midrule
Stribeck friction ($w^f$) &
$F_{c,j}$, $F_{s,j}$, $\sigma_{v,j}$ ($v_{s,j}$ fixed) &
$\times 2$ \\
Backlash ($w^b$) &
$B_j$ &
$\times 15$ \\
Compliance ($w^c$) &
$K_j=K_{j,0}/w^c_j$, $D_j=D_{j,0}/\sqrt{w^c_j}$
($M_j^{\mathrm{eff}}$ fixed) &
$K \div 5$, $D \div \sqrt{5}$ \\
Gravity ($w^g$) &
$\beta_j = w^g_j \beta_{j,0}$ &
$\times 4$ \\
\bottomrule
\end{tabular}
\caption{Full setting of each severity factor of \Eq{wear}. $w^f_j$, $w^b_j$, and $w^c_j$ model wear, and $w^g_j$ models payload mismatch.}
\label{tab:severity}
\end{table}

\subsection{Noise Injection Details}
\label{sec:supp_injection}

The VLA acts at $20$\,Hz, and the simulator advances each control step through $25$ sub-steps of $2$\,ms.
The operational space controller recomputes the joint torques $\tau_j$ at every sub-step, and all noise components are recomputed at the same rate, which is the time step $t$ of \Eq{backlash} and the step of the online integration of \Eq{compliance}.
The torque-level components are added to the controller torque before it is written to the actuator, and the sum is saturated at the joint torque limit.
The position-level components are implemented as kinematic offsets to the simulator's joint-position state before each physics integration sub-step.
Immediately before each integration sub-step, we add the change in the offset
$\eta^{\text{back}}_j + \eta^{\text{comp}}_j$ since the previous sub-step, so that the joint coordinate carries the current offset rather than an accumulated one.
The two layers are coupled only through compliance, whose load torque $\tau^{\text{load}}_j$ in \Eq{compliance} is the perturbed torque, whereas backlash depends solely on the motor-angle trajectory.
No noise is added to the observations; images and proprioception are both obtained from the resulting simulator state.
Because the joints are coupled through the mass matrix, a perturbation at one joint also changes the motion of the others.
The residual a policy observes is therefore not the injected noise itself but its effect through the arm's dynamics and kinematics at the current configuration.
In contrast, perturbations applied directly to end-effector commands, as in prior benchmarks~\citep{zhang2025robustvla, guo2025robustness}, do not represent the joint-level sources of these errors.

\subsection{Settings of Each Noise Component}
\label{sec:supp_isolated}
Each setting of Table~\ref{tab:isolated} activates a single noise component across all joints while disabling the other three, which is the setting used in Table~\ref{tab:unit}.
Table~\ref{tab:isolated} lists the parameter values for each component, separately for joints J1--J4 and J5--J7.

\begin{table}[!ht]
\centering
\small
\setlength{\tabcolsep}{6pt}
\begin{tabular}{@{}llL{7.2cm}@{}}
\toprule
Component & Setting & Injected values (J1--J4 / J5--J7) \\
\midrule
Stribeck   & friction $\times 2$ &
$F_c\,{=}\,4.350/0.960$\,Nm,
$F_s\,{=}\,6.960/1.440$\,Nm,
$\sigma_v\,{=}\,0.696/0.096$\,Nm$\cdot$s/rad,
$v_s\,{=}\,0.15/0.10$\,rad/s \\[2pt]
Backlash   & $B \times 15$ &
$B\,{=}\,0.01305/0.01950$\,rad
($0.748^{\circ}/1.117^{\circ}$) \\[2pt]
Compliance & $K \div 5$, $D \div \sqrt{5}$ &
$K\,{=}\,1200/800$\,Nm/rad,
$D\,{=}\,40.58/17.71$\,Nm$\cdot$s/rad,
$M^{\mathrm{eff}}\,{=}\,0.70/0.20$\,kg$\cdot$m$^2$
($\zeta = 0.70$) \\[2pt]
Gravity    & $\beta \times 4$ &
$\beta = [0.20,\, 0.40,\, 0.24,\, 0.40,\, 0.32,\, 0.32,\, 0.20]$ \\[2pt]
\bottomrule
\end{tabular}
\caption{Settings of noise components. Values are reported as J1--J4 / J5--J7 unless
stated otherwise.}
\label{tab:isolated}
\end{table}

\subsection{Deployment Scenarios}
\label{sec:supp_composite}
The deployment scenarios of \S\ref{sec:scenarios} combine the noise components as summarized in Table~\ref{tab:scenarios}.
Table~\ref{tab:composite} lists the components whose severity is increased in each scenario and their resulting parameter values.
The two Thermal Drift scenarios ramp one factor linearly over the episode from the mild setting to the full setting. The Stribeck variant ramps $w^f_j$ from $1.5$ to $2$ and the Backlash variant ramps $w^b_j$ from $5$ to $15$.
In Heavy Payload, $w^g_j$ and $w^c_j$ take the full setting on every joint while the gripper holds an object and stay nominal otherwise, and $w^f_j = w^b_j = 1$ throughout.
The gate turns on when the simulator's grasp check detects a held object and turns off three control steps after release, with a linear ramp-in of $0.1$\,s.
In the three Aged Joint cases, the worn joints are driven to the full setting while the remaining joints stay nominal, with the wear factors applied to all joints (Uniform), to J1--J3 (Shoulder), or to J4 (Elbow).
Unless stated otherwise, components not listed in a scenario retain their
nominal parameters from Table~\ref{tab:base_params}.
The Clean condition disables all noise components.

\begin{table}[!ht]
\centering
\small
\setlength{\tabcolsep}{4pt}
\begin{tabular}{@{}p{3.2cm} p{2.9cm} L{6.5cm}@{}}
\toprule
Scenario & Components with increased severity & Injected values \\
\midrule
Thermal Drift-Stribeck & Stribeck (ramp) &
$F_c$, $F_s$, and $\sigma_v$ ramp $\times 1.5\!\to\!\times 2.0$ over the episode:
$F_c\,{=}\,3.26\!\to\!4.35$ / $0.72\!\to\!0.96$\,Nm,
$F_s\,{=}\,5.22\!\to\!6.96$ / $1.08\!\to\!1.44$\,Nm,
$\sigma_v\,{=}\,0.522\!\to\!0.696$ / $0.072\!\to\!0.096$\,Nm$\cdot$s/rad.\\[3pt]
Thermal Drift-Backlash & Backlash (ramp) &
$B$ ramps $\times 5\!\to\!\times 15$: $0.00435\!\to\!0.01305$ / $0.0065\!\to\!0.0195$\,rad. \\[3pt]
Heavy Payload & Compliance + Gravity (while grasped) &
$K\,{=}\,1200/800$\,Nm/rad,
$D\,{=}\,40.58/17.71$\,Nm$\cdot$s/rad,
$M^{\mathrm{eff}}\,{=}\,0.70/0.20$\,kg$\cdot$m$^2$;
$\beta_j$ (J1--J7): 0.20, 0.40, 0.24, 0.40, 0.32, 0.32, 0.20. \\[3pt]
Aged Transmission & Stribeck + Backlash &
all joints: $F_c\,{=}\,4.35/0.96$\,Nm;
$F_s\,{=}\,6.96/1.44$\,Nm;
$\sigma_v\,{=}\,0.696/0.096$\,Nm$\cdot$s/rad;
$B\,{=}\,0.01305/0.01950$\,rad. \\[3pt]
Aged Joint-Uniform & Stribeck + Backlash + Compliance &
all joints: the above Stribeck and backlash parameters plus
$K\,{=}\,1200/800$\,Nm/rad,
$D\,{=}\,40.58/17.71$\,Nm$\cdot$s/rad. \\[3pt]
Aged Joint-Shoulder & Stribeck + Backlash + Compliance &
J1--J3 at the full setting; J4--J7 nominal. \\[3pt]
Aged Joint-Elbow & Stribeck + Backlash + Compliance &
J4 at the full setting; the other joints nominal. \\
\bottomrule
\end{tabular}
\caption{Deployment scenarios. Values are reported as J1--J4 / J5--J7 unless stated otherwise. ``Full setting'' applies the factors of Table~\ref{tab:severity}, and ``nominal'' denotes the values of Table~\ref{tab:base_params}.}
\label{tab:composite}
\end{table}

\section{Implementation Details}
\label{sec:supp_impl}

\subsection{Evaluation Protocol}
\label{sec:supp_protocol}
We evaluate on four LIBERO suites: LIBERO-Spatial, LIBERO-Object, LIBERO-Goal, and LIBERO-Long, and report the success rate averaged over tasks. Every scenario in Tables~\ref{tab:isolated} and~\ref{tab:composite} is evaluated on all four suites with $20$ episodes per task and three seeds, which vary the policy's sampling noise while the initial states are shared, and we report the mean over seeds.
For each backbone~\citep{black2024pi_0,black2025pi_}, all methods start from the same pretrained checkpoint and use the same action horizon
($K = 50$, $n_{\text{exec}} = 5$), per-task initial states, and scenario parameters.
The hidden noise states, including the backlash link position and compliance deflection, are reset at the start of each episode.
Because the scenarios evolve in closed loop, their realized execution errors depend on each policy's state trajectory.

\subsection{Configuration of Our Method}
\label{sec:supp_method}
We freeze each backbone and attach a rank-$4$ LoRA to its action expert,
updating only the LoRA parameters with Adam (learning rate $10^{-6}$, anchor
weight $\lambda_{\text{anc}} = 10^{-4}$; \Eq{flow-matching}).
The policy predicts an action chunk of horizon $K = 50$, of which the robot
executes the first $n_{\text{exec}} = 5$ steps before the next prediction.
After each chunk, we compute the residual of \Eq{eta-obs} for the executed
steps and store it in a FIFO buffer containing the $32$ most recent chunks.
Once the buffer is full, we update the LoRA every $8$ chunks with one Adam step
on a batch of $4$ sampled chunks.
Since only $5$ of the chunk's $K{=}50$ predicted steps are executed, only those
steps receive residual correction, while the remaining actions retain their
original policy values in the target chunk.

In simulation, we carry the LoRA parameters and Adam state across all episodes
of the ten tasks in a suite, starting from $\phi_0$.
We evaluate tasks in LIBERO index order and clear the buffer at each task
boundary, so each update uses feedback from the current task.
We re-initialize the LoRA parameters, Adam state, and buffer for each suite and
scenario.

\subsection{Action and Observation Space}
\label{sec:supp_actspace}
The policy outputs a delta action $a = (\Delta p, \Delta \omega, g)$, where $\Delta p \in \mathbb{R}^3$ is the world-frame end-effector translation delta, $\Delta \omega \in \mathbb{R}^3$ the axis-angle rotation delta, and $g \in \{-1, +1\}$ the binary gripper command. In our policy implementation, this 7-dimensional action is represented as an 8-dimensional vector in $[-1, 1]^8$ with one unused channel.
The controller maps the normalized action to physical units through
$(c_p, c_r) = (0.05\,\text{m},\ 0.5\,\text{rad})$. The proprioceptive observation is
$s = (p,\ \mathrm{axisangle}(R),\ g^{\text{qpos}}) \in \mathbb{R}^8$, together with two
$224 \times 224$ RGB views (third-person and wrist) and the language instruction.
To obtain the achieved delta $\Delta x_{t,i}$ of \Eq{eta-obs}, we use Euclidean
subtraction for translation and the $SO(3)$ logarithm map for rotation:
\begin{equation}
\Delta x_{t,i} = \Big(
  \tfrac{p_{t,i+1} - p_{t,i}}{c_p},\ 
  \tfrac{\mathrm{Log}_{SO(3)}\!\big(R_{t,i+1} R_{t,i}^{-1}\big)}{c_r},\ 
  g^{\text{cmd}}_{t,i}
\Big),
\end{equation}
where $c_p$ and $c_r$ rescale the deltas to the normalized action range and the log
map is taken along the short path (the rotation of smallest angle). We set the
gripper entry to the commanded value $g^{\text{cmd}}$, which excludes the gripper from
the residual since its command is binary while its proprioceptive state is continuous, so a
residual on it would not be meaningful.
Any remaining padding dimensions are retained from the original policy output and receive no residual correction.

\subsection{LoRA and Flow-Matching Optimization}
\label{sec:supp_lora}
The rank-$4$ LoRA is attached to all linear projections of the $300$M-parameter Gemma
action expert (query, key, value, and output, and the MLP gate, up, and down
projections), giving $2{,}764{,}800$ trainable parameters, $0.085\%$ of the $\pi_0$
backbone's $3.24$\,B. Following \citet{hu2022lora}, we initialize
$A \sim \mathcal{N}(0, 0.01^2)$ and $B = 0$, so $AB = 0$ initially and the adapted
policy matches the base policy before any update. The PaliGemma vision-language
backbone is frozen.
We minimize the flow-matching loss of \Eq{flow-matching} with Adam, drawing one Monte-Carlo
$(u, \epsilon)$ sample per update with $u \sim \mathrm{Uniform}(0, 1)$ and
$\epsilon \sim \mathcal{N}(0, I)$. We set $(\beta_1, \beta_2) = (0.9, 0.999)$, use no
weight decay and no learning-rate schedule, and clip gradients to a global
$\ell_2$-norm of $1.0$.

\subsection{Computational Setup}
\label{sec:supp_compute}
All evaluations run on a single NVIDIA A6000 ($48$\,GB) GPU. The codebase builds on
JAX 0.5.0 with Flax-NNX, robosuite 1.4.1, and MuJoCo 3.7.0 within the LIBERO
benchmark. 
The control loop runs at $20$\,Hz ($50$\,ms per step), and MuJoCo integrates the physics at $500$\,Hz ($25$ sub-steps of $\Delta t = 2$\,ms per control
step). 
Per chunk, multi-step flow-matching inference takes $\sim$$80$\,ms and one LoRA
Adam step (batch of $4$) takes $\sim$$50$\,ms.
The maximum episode
length is $220$ steps for LIBERO-Spatial, $280$ for LIBERO-Object, $300$ for
LIBERO-Goal, and $520$ for LIBERO-Long, corresponding to $11$ to $26$\,s of simulated
time.

\subsection{Real-Robot Setup}
\label{sec:supp_real}

\paragraph{Platform and demonstrations.}
Each AgileX Piper arm has six revolute joints and a parallel gripper, so the state and action are $7$-dimensional.
The policy observes two RGB views from base and wrist cameras, the joint state, and the language instruction, and the control loop runs at $30$\,Hz.
On each arm, we collect $30$ teleoperated demonstrations for each of the four tasks, recording the commanded absolute joint targets as actions.
Each base policy is fine-tuned with the standard flow-matching objective on $50$-step action chunks for $20{,}000$ steps, representing joint actions as displacements from the joint state at the start of each chunk.

\paragraph{Real-robot scheduling.}
The Piper control loop executes $n_{\text{exec}}=10$ actions per prediction.
After each execution window, the client computes the residuals and asynchronously sends feedback to the policy server through a separate WebSocket connection.
The server schedules an update every eight feedback chunks.
Prediction requests and adaptation updates are handled asynchronously, and completed updates affect subsequent predictions without modifying already generated commands.
The server uses the same FIFO buffer, update interval, and optimization settings as in \S\ref{sec:supp_method}, unless stated otherwise.

\paragraph{Evaluation state and carry-over.}
For each arm, task, and backbone, evaluation starts from the corresponding fine-tuned base policy with a zero-effect LoRA initialization, a fresh Adam
state, and an empty feedback buffer.
We retain the LoRA parameters, Adam state, and feedback buffer across the $25$ consecutive evaluation episodes of the same task, and re-initialize all three
states at each task boundary.
No episodes are reserved solely for adaptation: the reported evaluation episodes also populate the feedback buffer, and online updates begin once the buffer is full.

\paragraph{Residual and pseudo-target.}
On the physical robots, we compute the residual per joint in radians as
\begin{equation}
\eta^{\text{obs}}_{t,i}
=
q_{t,i+1}-q^{\text{ref}}_{t,i},
\end{equation}
where $q^{\text{ref}}_{t,i}$ is the joint target predicted by the policy before
any controller-side correction, and $q_{t,i+1}$ is the joint state measured at
the next control step.
The pseudo-target is
$a^{\text{target}}_{t,i}=q^{\text{ref}}_{t,i}-\eta^{\text{obs}}_{t,i}$, as in
\Eq{corrected}.
We exclude the gripper dimension from the residual and retain its original
policy target.
The residual and pseudo-target are formed in absolute joint coordinates.
Before computing the flow-matching loss, we convert the pseudo-target to the representation used in fine-tuning: we subtract the joint state at the start of the chunk from the six joint coordinates, keep the gripper target unchanged, and apply the checkpoint's action normalization. 

\paragraph{DOB implementation.} For the DOB baseline, we use an identity nominal model for each joint servo and estimate the equivalent input disturbance with a first-order Q-filter of bandwidth $\omega_Q = 2.0\,\mathrm{s}^{-1}$.
At each control step, the DOB clips the estimate $z_t$ to $\hat{d}_t = \mathrm{clip}(z_t, -d_{\max}, d_{\max})$ with $d_{\max} = 0.05$\,rad and sends the corrected target $q^{\text{DOB}}_t = q^{\text{ref}}_t - \hat{d}_t$ to the robot controller.
We update the estimate at 30\,Hz using the elapsed control interval, clamped to $[0.005, 0.2]$\,s, and reset the DOB state at the start of each episode.
For DOB+Ours, our method computes the residual against the policy target before the DOB correction, $\eta^{\text{obs}}_t = q_{t+1} - q^{\text{ref}}_t$.
As the DOB reduces the tracking error, this residual shrinks, so our method adapts only to the error that the DOB does not cancel.
Computing the residual against $q^{\text{DOB}}_t$ would include the DOB's own correction and could compensate twice.

\paragraph{OOD evaluation.}
We evaluate OOD generalization using $\pi_{0.5}$.
Each OOD evaluation starts from the corresponding fine-tuned $\pi_{0.5}$ policy with the LoRA parameters initialized to $\phi_0$, a fresh Adam state, and an empty feedback buffer.
We retain the adaptation state across the OOD episodes and object variants.

\section{Residual Magnitude across Scenarios}
\label{sec:residual-magnitude}

Table~\ref{tab:residual-magnitude} reports the normalized residual magnitude,
$\|\eta^{\text{obs}}\|_2 / \|a^{\text{policy}}\|_2$, measured on $\pi_0$
rollouts.
For each rollout, we compute the normalized residual over executed non-gripper actions
and report the mean across rollouts.
Even in the Clean setting, the residual is $23.4\%$, reflecting the nominal tracking mismatch.
Across the seven RoboStress deployment scenarios, it increases to
$38.7\%$--$56.3\%$, or $15.3$--$32.9$ percentage points above Clean, showing
that the modeled execution variations induce substantial additional deviations
from the policy commands.
On the two Piper arms, the residual is $32.9\%$ for Robot A and $35.0\%$ for
Robot B, confirming that command-execution mismatch also occurs during physical deployment.
The two groups of rows are not on a common scale.
The simulation rows use the end-effector residual of \Eq{eta-obs} in the normalized action units of \S\ref{sec:supp_actspace} at $20$\,Hz with $n_{\text{exec}}=5$, whereas the real-robot rows use the joint-space residual of \S\ref{sec:supp_real} in radians at $30$\,Hz with $n_{\text{exec}}=10$.
RoboStress complements these physical experiments by covering execution conditions that are difficult to realize or to vary systematically on physical hardware.

\begin{table}[h]
\centering
\setlength{\tabcolsep}{8pt}
\renewcommand{\arraystretch}{1.15}
\begin{tabular}{l c}
\toprule
Scenario & Residual (\%) \\
\midrule
\multicolumn{2}{l}{\textit{Simulation (RoboStress)}} \\
\midrule
Clean & 23.4 \\
Heavy Payload & 48.6 \\
Thermal Drift-Stribeck & 47.5 \\
Thermal Drift-Backlash & 46.1\\
Aged Transmission & 38.7 \\
Aged Joint-Uniform & 48.2 \\
Aged Joint-Shoulder & 56.3 \\
Aged Joint-Elbow & 47.0 \\
\midrule
\multicolumn{2}{l}{\textit{Real robots (AgileX Piper)}} \\
\midrule
Robot A & 32.9 \\
Robot B & 35.0 \\
\bottomrule
\end{tabular}
\caption{Residual magnitude across scenarios, computed as
$\|\eta^{\text{obs}}\|_2 / \|a^{\text{policy}}\|_2$ on $\pi_0$ rollouts and expressed
as a percentage. The simulation rows use the end-effector residual in normalized action units and the real-robot rows use the joint-space residual of \S\ref{sec:supp_real} in radians.}
\label{tab:residual-magnitude}
\end{table}

\section{Additional Results}
\label{sec:supp_results}

\subsection{Impact of the Anchor Loss}
The anchor term in the compensation objective penalizes deviations of the LoRA parameters from their initialization.
Table~\ref{tab:anchor-ablation} compares success rates with and without this term on $\pi_0$, averaged over the four LIBERO suites.
Removing the anchor lowers success in all seven deployment scenarios, reducing the average from $44.2\%$ to $41.8\%$ ($2.4$ percentage points).
The largest drop occurs under Heavy Payload ($4.6$ points), followed by Aged Transmission and Aged Joint-Shoulder ($3.7$ points each).
We therefore retain the anchor term to regularize deployment-time adaptation.

\begin{table}[h]
\centering
\setlength{\tabcolsep}{10pt}
\renewcommand{\arraystretch}{1.15}
\begin{tabular}{l c c}
\toprule
Scenario & w/o anchor & w/ anchor \\
\midrule
Heavy Payload & 38.2 & 42.8 \\
Thermal Drift-Stribeck & 50.4 & 50.6 \\
Thermal Drift-Backlash & 77.9 & 78.8 \\
Aged Transmission & 25.2 & 28.9 \\
Aged Joint-Uniform & 22.8 & 24.0 \\
Aged Joint-Shoulder & 35.5 & 39.2 \\
Aged Joint-Elbow & 42.8 & 44.9 \\
\midrule
Avg. & 41.8 & 44.2 \\
\bottomrule
\end{tabular}
\caption{Effect of the anchor loss on $\pi_0$.
Success rates (\%) per deployment scenario, averaged over the four LIBERO suites.}
\label{tab:anchor-ablation}
\end{table}

\subsection{Performance across Severity Levels}
\label{sec:supp_severity_levels}

\begin{figure}[t]
\centering
\includegraphics[width=\linewidth,height=0.9\textheight,keepaspectratio]{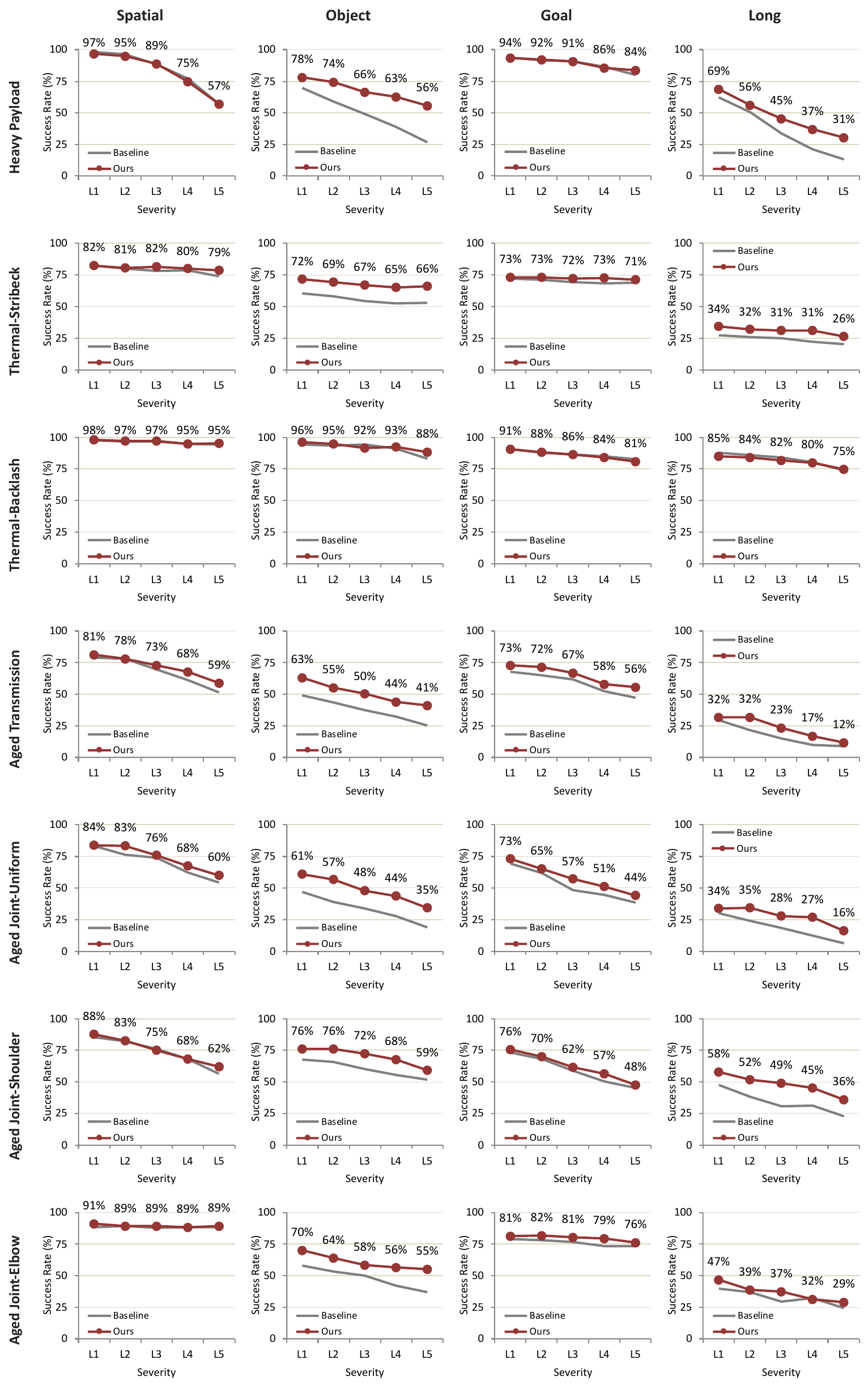}
\caption{Success rate (\%) of the base policy and self-compensating VLA with $\pi_{0.5}$ across five severity levels (L1 to L5) for each deployment scenario (rows) and LIBERO suite (columns) in RoboStress, averaged over the same three seeds as Table~\ref{tab:realistic}.}
\label{fig:supp_severity}
\end{figure}

To examine how performance changes with the strength of execution errors, we evaluate each deployment scenario at five severity levels, L1 to L5.
Each level moves every active component linearly from the \emph{mild} setting (L1) to the full setting of Table~\ref{tab:severity} (L5), $w_k = w_{L1} + \tfrac{k-1}{4}\,(w_{L5}-w_{L1})$ (Table~\ref{tab:supp_severity_levels}).
The payload factor $w^g_j$ follows the same rule, and for the two Thermal Drift scenarios the level sets the endpoint of the within-episode ramp while its onset stays fixed, so L1 applies the mild setting throughout the episode. In the Aged Joint scenarios only the worn joints move along the ladder; the remaining joints stay nominal.
L5 is the full setting used in Table~\ref{tab:realistic}. We use $\pi_{0.5}$ in this analysis.

\begin{table}[!ht]
\centering
\small
\setlength{\tabcolsep}{5pt}
\begin{tabular}{@{}l ccc cc cc@{}}
\toprule
 & \multicolumn{3}{c}{Wear factors} & \multicolumn{2}{c}{Heavy Payload}
 & \multicolumn{2}{c}{Thermal Drift ramp end} \\
\cmidrule(lr){2-4}\cmidrule(lr){5-6}\cmidrule(l){7-8}
Level & Stribeck $w^f$ & Backlash $w^b$ & Compliance $K\div$
      & $K\div$ & $\beta\times$ & Stribeck & Backlash \\
\midrule
L1 (mild) & 1.5   & 5    & 2    & 2    & 2   & 1.5$\to$1.5   & 5$\to$5 \\
L2          & 1.625 & 7.5  & 2.75 & 2.75 & 2.5 & 1.5$\to$1.625 & 5$\to$7.5 \\
L3          & 1.75  & 10   & 3.5  & 3.5  & 3   & 1.5$\to$1.75  & 5$\to$10 \\
L4          & 1.875 & 12.5 & 4.25 & 4.25 & 3.5 & 1.5$\to$1.875 & 5$\to$12.5 \\
L5 (full)& 2     & 15   & 5    & 5    & 4   & 1.5$\to$2     & 5$\to$15 \\
\bottomrule
\end{tabular}
\caption{Severity levels. Each column moves linearly from L1 to L5, and L5 is the setting of Table~\ref{tab:realistic}. Compliance divides the stiffness by the listed factor and the damping by its square root ($K_j=K_{j,0}/w^c_j$, $D_j=D_{j,0}/\sqrt{w^c_j}$), which keeps $\zeta=0.70$ at every level. Wear factors apply to the worn joints of each scenario, that is, all joints for Aged Transmission and Aged Joint-Uniform, J1--J3 for Shoulder, and J4 for Elbow. Thermal Drift entries give the ramp as onset$\to$end multiplier.}
\label{tab:supp_severity_levels}
\end{table}

Figure~\ref{fig:supp_severity} shows the results.
Success generally decreases as severity increases for both methods, and the rate of decrease depends on the scenario.
Heavy Payload is the most sensitive as the success rate of self-compensating VLA averaged over the four suites drops from 84.3\% at L1 to 56.8\% at L5.
Thermal Drift-Stribeck and Thermal Drift-Backlash are the least sensitive, dropping from 65.2\% to 60.5\% and from 92.4\% to 84.8\%, respectively, and success on LIBERO-Spatial under Aged Joint-Elbow remains around 90\% at every level.
Across suites, LIBERO-Long is the most affected, falling to about two-thirds of its L1 success rate, since errors accumulate over its longer horizon.
Self-compensating VLA generally matches or outperforms the base policy at nearly every severity level. The gap widens with severity on Heavy Payload and the Aged Joint scenarios, most notably on LIBERO-Object and LIBERO-Long, while it stays roughly constant on the Thermal Drift scenarios.

\end{document}